\documentclass[runningheads]{llncs}
\usepackage{graphicx}
\usepackage{amsmath}
\usepackage{bm}
\usepackage{amssymb}
\begin{document}
\title{Automated Enriched Medical Concept Generation for Chest X-ray Images}
%
%\titlerunning{Leveraging Clinical Reports for Concept Detection in Chest X-ray Images}
\author{Aydan Gasimova}%\inst{1} %\and
%Second Author\inst{2,3}\orcidID{1111-2222-3333-4444} \and
%Third Author\inst{3}\orcidID{2222--3333-4444-5555}
\institute{Imperial College London} %\and
%\email{a.gasimova16@imperial.ac.uk}\\
\maketitle              % typeset the header of the contribution
\begin{abstract}
Decision support tools that rely on supervised learning require large amounts of expert annotations. Using past radiological reports obtained from hospital archiving systems has many advantages as training data above manual single-class labels: they are expert annotations available in large quantities, covering a population-representative variety of pathologies, and they provide additional context to pathology diagnoses, such as anatomical location and severity. Learning to auto-generate such reports from images present many challenges such as the difficulty in representing and generating long, unstructured textual information, accounting for spelling errors and repetition/redundancy, and the inconsistency across different annotators. We therefore propose to first learn visually-informative medical concepts from raw reports, and, using the concept predictions as image annotations, learn to auto-generate structured reports directly from images. We validate our approach on the OpenI~\cite{openi} chest x-ray dataset, which consists of frontal and lateral views of chest x-ray images, their corresponding raw textual reports and manual medical subject heading (MeSH\textsuperscript{\textregistered}) annotations made by radiologists.
\keywords{nlp  \and medical imaging \and deep learning}
\end{abstract}

\section{Introduction}
Radiologists are faced daily with the very time-consuming and repetitive task of looking at hundreds of radiography images and writing up radiological reports. The fast turn-arounds they are expected to produce leads to fatigue that can negatively affect diagnostic accuracy~\cite{stec2018systematic}. Supervised learning for automated pathology detection from images has the potential for clinical-decision support, however, such image segmentation and classification learning tasks require detailed annotations covering a large distribution of input data for the algorithms to be able to make robust predictions. Such annotations must be made by qualified radiologists, which, for the detail and breadth of annotation required, will be an equally if not more time-consuming task than manually creating the reports. In addition, classification and semantic segmentation tasks only solve for the prediction of presence of pathologies, and not the generation of reports which contain additional information such as severity, location, and absence of pathologies. 

Recently, we have seen supervised learning approaches that aim to take advantage of past radiological exams containing reports in order to either auto-generate the reports~\cite{shin2016learning,jing2017automatic,zhang2017tandemnet}, or to assist in classification tasks~\cite{shin2016interleaved,wang2017chestx,wang2017unsupervised,zhang2017mdnet,yan2019holistic}. The noise present in medical reports in addition to the presence of non-visually significant information, such as the negation of pathologies, make it difficult to learn from them directly as done in natural image captioning frameworks. Additionally, high recall/precision of pathologies is more crucial in the medical domain where the risk of mis-labelling is much higher.

We therefore propose to use a limited number of manual medical concept annotations in order to first learn to extract them from raw reports, and then take advantage of the model predictions as image annotations, thus providing a method for augmenting an image-annotation dataset. We then demonstrate how these image-concept annotations can be learned through sequence models conditioned on image features, and generate a more readable context for the diagnosis that can be used as part of a clinical decision support system, thus greatly alleviating the burden on radiologists. Our approach can be summarised in the following steps:

\begin{enumerate}
    \item We propose a network that learns to extract visually-significant medical concepts from raw reports. To our knowledge, this is a first attempt to that goes beyond simple pathology detection to include concepts such as anatomical position and severity.
    \item We explore several sequence-learning networks that aim to condition the sequence generation process on image features in order to learn to auto-generate structured reports from radiological images.
    \item We use the predictions made by the structured report generation process in step 1 to demonstrate how they can be used to create an image-report training set for step 2.
\end{enumerate}

\section{Related Work}
\subsection{Data-mining Image Labels}
There are two common approaches to extracting image labels from raw reports: statistical and tool based. Radiological text mapping tools such as DNorm~\cite{leaman2015challenges} and MetaMap~\cite{aronson2001effective} have been used to extract labels for multi-label classification~\cite{yan2019holistic} and in weakly supervised localisation learning frameworks~\cite{wang2017chestx}. However, other biological concepts in the reports, such as location, severity, and other visually descriptive features of the pathology are not taken advantage of. Unsupervised, statistical methods such as latent Dirichlet allocation \cite{shin2016interleaved} and clustering~\cite{wang2017unsupervised} have been used to implicitly define topics and cluster groups containing key words and propose classification into these topics and groups. These approaches are heavily dependent on the number of topics/groups providing the lowest perplexity score, which can be a range of values. In addition, these are not generative models, therefore reports can only be selected based on nearest-neighbour methods. To this end, we propose instead to learn to generate reports comprised of medical concept from images, in a similar style to natural image captioning.
 
\subsection{Radiology Report Generation}
Closest to our work, Shin et al.\ proposed a cascaded learning framework to auto-generate MeSH annotations from chest X-rays~\cite{shin2016learning} whereby image embeddings are first extracted from a pre-trained classification network, and then used to initialise a sequence prediction model to auto-generate MeSH sequences. Zhang et al.~\cite{zhang2017mdnet,zhang2017tandemnet} leverage manually created structured reports in a dual-attention framework to improve features used for classifying histopathology images and to provide interpretability to the classification. The reports used in both cases are far more structured than their raw counterparts and so this approach cannot be directly translated to hospital data. Training on raw hospital reports, Jing et al.~\cite{jing2017automatic} demonstrated how they can be generated by first training a multi-label CNN on the images and the Medical Text Indexer (MTI) tags identified in the original raw reports of the Openi chest x-ray dataset. However, reports can be very long and heterogeneous, and the authors do not evaluate the model's ability to determine whether visually and clinically-relevant medical concepts have been identified. To address the challenges of learning from raw reports directly, we first learn to generate structured reports made up of only visually-significant medical concepts that correspond directly to features seen in the images. Being shorter and vocabulary-controlled, the generation process is easier to evaluate for correct identification of pathologies.

\section{Method}
\subsection{Enriched Concept Extraction from Raw Reports}
We approach learning structured reports from raw textual reports as a multi-label classification task since the vocabulary of MeSH terms is consistent across annotators, and limited. We modify the shallow CNN first introduced by Kim~\cite{kim2014convolutional} for multi-class text classification and later adapted for multi-label text classification by Liu at al.~\cite{liu2017deep} by introducing a learn-able embedding layer as we do not have the advantage of pre-trained word embeddings for medical text, and by introducing dropout followed by a fully-connected layer to each convolutional output prior to the concatenation to aid regularisation.

Let $\bm{x}_i\in\mathbb{R}^d$ be the d-dimensional word vector for the $i$-th word of report $\bm{p}$. The textual report is thus represented as a concatenation of word embeddings: $\bm{p} = [\bm{x}_0, ... \bm{x}_i, ... \bm{x}_M] \in\mathbb{R}^{M\times d}$ where $M$ is the maximum length of the reports. A filter $\bm{m}\in\mathbb{R}^{hd}$ is convolved with a window of $h$ words to produce a new feature $c_i$:

\begin{equation}
 c_j = f(\bm{m}*\bm{x}_{i:i+h-1}+b)
\end{equation}

where $f$ is a non-linear activation function and $b$ is a bias term. The filter is applied consecutively to every $h$-word window in the sentence, resulting in a feature map $\bm{c}=[c_0, .... c_i, ... c_{M-h+1}] \in\mathbb{R}^{M-h+1}$. Max-over-time pooling~\cite{collobert2011natural} is applied over each feature map to capture the most important feature $\hat{c}=max(\bm{c})$. In this way we apply many filter operations with varying window widths in order to obtain multiple features that are able to capture semantic information of reports with varying word lengths. We use the sigmoid activation function as we require an independent prediction for each class and train by minimising the multi-class sigmoid cross-entropy (SCE) loss. In addition, we add terms to balance maximising the true positive class prediction with true negative class predictions as the positive label space is very sparse:

\begin{equation}
 \begin{split}
 \widehat{SCE}_i = &- \lambda_1 \sum_{j=1}^K \left(  y_{j}log(f(s_{ij})) + (1-y_{j}log(1-f(s_{j})) \right) \\
 &- \lambda_2 \sum_{j=1}^K y_{j}f(s_{j})/ \sum_{j=1}^K (y_{j}f(s_{j}) + y_{j}(1-f(s_{j}))) \\
 &- \lambda_3 \sum_{j=1}^K (1-y_{j})(1-f(s_{j})) / \sum_{j=1}^K ((1-y_{j})(1-f(s_{j})) + (1-y_{j})(f(s_{j})))
\end{split}
\end{equation}
%\begin{equation}
%\widehat{SCE} = \frac{1}{N} \sum_{i=1}^N \widehat{SCE_i}
%\end{equation}
where $K$ is the number of classes, $y_{j}$ is presence/absence of class label $j$ for instance $i$, $f(s_{j})$ is the prediction for instance $i$ on label $j$ made through a sigmoid activation:

\begin{equation}
 f(s_i) = \frac{1}{1+e^{-s_i}}
\end{equation}

The weights of each loss term, $\lambda_1, \lambda_2, \lambda_3$ are non-negative, sum to 1 and chosen through cross-validation. Finally, the modified SCE loss is averaged over batches.
%The full architecture is displayed in Figure~\ref{fig:text_cnn_single_output}.

\subsection{Report Generation from Images}
Given that it is possible to learn structured report outputs from raw reports, we propose a method of learning to auto-generate structured reports directly from images. We explore multiple ways of conditioning the MeSH sequence learning on the image embeddings that aims to maintain the dependency between the word generation process and the image embedding at every time-step. The MeSH sequence is modelled using an RNN, specifically the Long Short-Term Memory (LSTM) implementation proposed in~\cite{hochreiter1997long}. Each LSTM unit has three sigmoid gates to control the internal state: `input', `output' and `forget'. At each time step, the gates control how much of the previous time steps is propagated through to determine the output. For an input word sequence $\{x_1, \dots, x_n \}$ where $x_i \in \mathbb{R}^d$, the internal hidden state $h_t \in \mathbb{R}^{h}$ and memory state $m_t \in \mathbb{R}^{m}$ are updated as follows:

\begin{equation}
\begin{split}
h_t &= f_t \odot h_{t-1} + i_t \odot \tanh(W^{(hx)}x_t + W^{(hm)}m_{t-1}) \\
m_t &= o_t \odot \tanh(h_t)
\end{split}
\end{equation}

where $ x_t \in \mathbb{R}^D$ is the word embedding, $W^{(hx)}$ and $W^{(hm)}$ are the trainable weight parameters, and $i_t$, $o_t$ and $f_t$ are the input, output and forget gates respectively. Bias terms are left out for readability.

The image embedding, $\bm{im}_i=\text{CNN}(I)$ where $\bm{im}_i \in \mathbb{R}^g$ is extracted from the final spatial-average pooling layer of the pre-trained CNN. We explore three ways of conditioning the sequence learning process:

\begin{enumerate}
\item RNN0: The image embedding is projected into the same embedding space as the word embeddings via a dense transition layer: $\bm{im} = \text{relu}(W^{(dg)}\text{CNN}(I))$. The image embedding is concatenated with the word sequence and thus treated as the first `word' in the MeSH sequence. 
\item RNN1: The image embedding is projected via a dense transition layer into a fixed embedding width and combined with the output of the recurrent layer through either concatenation or summation operation, and passed to the decoder $dec$:
\begin{equation}
\bm{dec}_t = \text{relu}(W^{(z)}(o_t*\text{relu}(W^{(dg)}\text{CNN}(I)))
\end{equation}

where $*$ represents concatenation or summation and $W^z$ are the weights of the decoder.

\item RNN2: The image embedding is projected via a dense transition layer into a fixed embedding width and combined with the input of the recurrent layer through either concatenation or summation operation, and passed to the encoder $enc$:
\begin{equation}
\bm{enc}_t = \text{relu}(W^{(a)}(x_t*\text{relu}(W^{(dg)}\text{CNN}(I)))
\end{equation}
where $W^a$ are the weights of the encoder.
\end{enumerate}

\begin{figure}
\includegraphics[width=\textwidth]{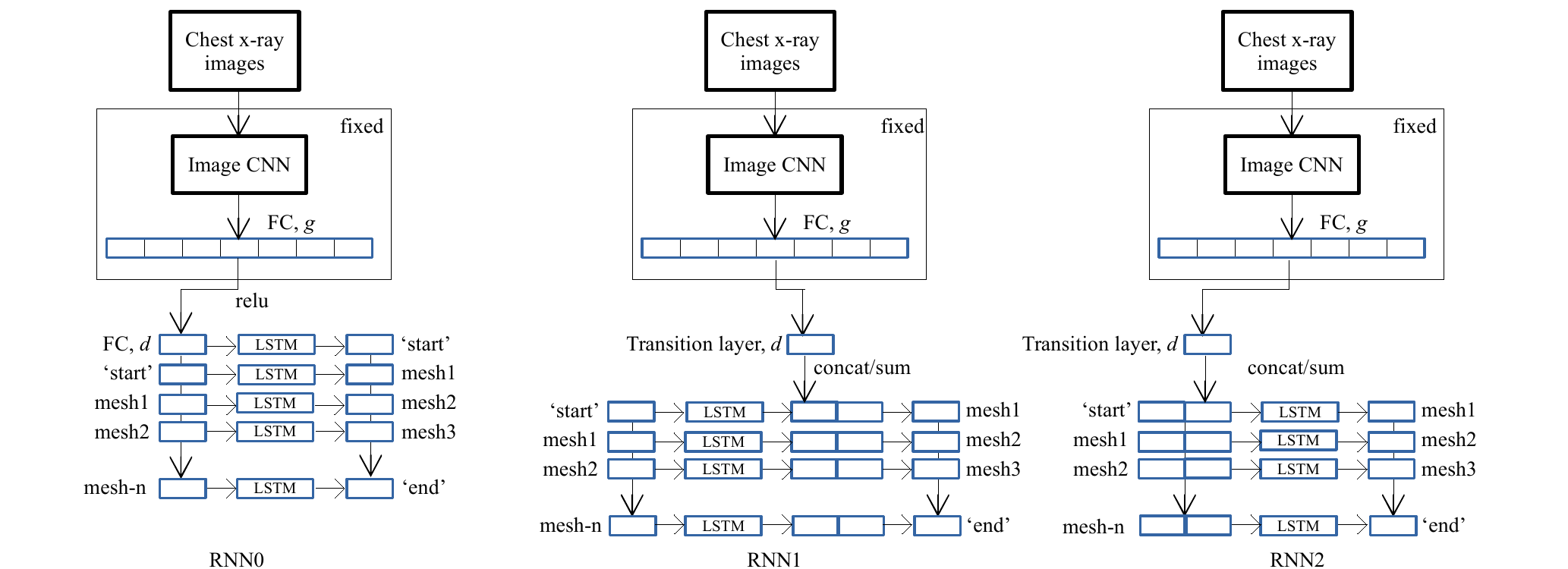}
\caption{Image-MeSH sequence learning model architectures.} 
\label{fig:seq_models}
\end{figure}

The model architectures are illustrated in Figure~\ref{fig:seq_models}. For all models, the decoder outputs are passed to the prediction layer $s(t) = f(W^{T}x_t)$ where $f$ is the softmax function. The models are all trained by minimising the cross-entropy loss  between the output and true sequence:

\begin{equation}
L(S,I) = -\sum_{t=0}^{T} \log p(P_t = T_t | \text{CNN(I)}, P_0 \ldots P_{t-1})
\end{equation}

where $p$ is the probability that the predicted word $P_t$ equals the true word $T_t$ at time step $t$ given image features $\text{CNN}(I)$ and previous words $P_0 \ldots P_{t-1}$, and $T$ is the LSTM sequence length.

\section{Experiments}
\subsection{Dataset}
We evaluated our models on the OpenI~\cite{openi} Indiana U. Chest X-ray Collection. This dataset consists of 7,470 frontal and lateral chest x-ray images and 3,955 associated radiological reports from the hospital's picture archiving systems. They have all been fully anonymised to remove patient names. In addition to the raw text reports, each exam has MeSH annotations made by qualified radiologists. MeSH annotations are (with some exceptions) formatted as [pathology/description,... pathology/description] where \emph{description} is a combination of \emph{anatomy/position/severity}. The number of captions per image is on average 2.33, with an average of 2.68 MeSH terms per caption.
\subsubsection{Preprocessing} This involved lower-casing, punctuation and non-alpha-numeric character removal from reports and MeSH. We limit the MeSH annotations to just one \emph{pathology/description} pair by selecting the caption with the most common pathology. Additionally, as the negation of pathologies was generally standard across reports, we performed negation removal using regex. Finally, the text reports were cropped/padded to 32 words based on the average of 20.23 +1 std of 11.9. MeSH captions were cropped/padded to length 5 based on average+1 std. Empty reports were removed. This resulted in 3,023 unique report-MeSH term pairs, of which 300 were randomly selected for validation and 300 for test.

\subsection{Experimental Settings}
We first investigate whether the structured reports can be learned from raw reports by creating a sub-set of size=1000 of the MeSH annotated reports, and training the text CNN on the report-MeSH pairs in the sub-set. The trained text CNN is then used to make MeSH prediction on the remaining set of raw reports, and these (together with the gold-standard annotated sub-set from the previous step) are used to train the image-MeSH sequence model. We compare this to training the image-MeSH sequence model on the entire gold-standard annotated set of 3,023.

\subsubsection{Text CNN}
For the text CNN model, we use rectified linear units as activation function on the convolutional layers, one-dimensional convolutional filters of width 3,4,5 with 512 feature maps for each filter, dropout rate~\cite{dahl2013improving} of p=0.5, with 254 hidden units for the dense layer, and with $\lambda_1=0.5, \lambda_2=0.2, \lambda_3=0.3$ for the loss terms. The model was trained through batch backpropagation, batch size=128 and using Adam optimisation~\cite{kingma2014adam} with learning rate=0.001 for 100 epochs with early-stopping. To compensate for the class imbalance of `normal' vs.\ diseased cases, we select batches with uniform distribution over the classes, augmenting the instances by sentence-shuffling.
\subsubsection{Sequence Models}
Image embeddings are extracted from the last average pooling layers of Vgg16~\cite{simonyan2014very} and Resnet50~\cite{he2016deep} models, pre-trained on ImageNet~\cite{deng2009imagenet} to extract $\bm{im} \in \mathbb{R}^{4096}$ and $\bm{im} \in \mathbb{R}^{2048}$ respectively. For RNN0, the joint image-word embedding dim is set to 2048 for the Vgg input, and 1024 for Resnet. For RNN1 and RNN2, the dense transition layer dimension is set to 1024. For all the sequence models, the LSTM hidden state is set to dim 512, and the LSTM units are unrolled up to 6 time steps (1 for the start token, and 5 for MeSH sequence). All models are trained with batch size 128, using Adam optimisation~\cite{kingma2014}, learning rate=0.001 and early-stopping.

\begin{table}
\caption{Text CNN classification metrics for sub-sampled and full gold-standard annotated data. Metrics reported on test data.}
\begin{center}
\begin{tabular}{|l|c|ccc|ccc|}
\hline
\multicolumn{8}{|c|}{All Classes} \\
\hline
Training sample size & Acc. & R & R-OC & R-OS & P & P-OC & P-OS \\
\hline
1000 & 98.26 & 67.82 & 40.49 & 65.79 & 70.15 & 45.15 & 67.18 \\
3023 (all) & 99.48 & 92.07 & 84.50 & 91.77 & 89.90 & 84.82 & 89.47 \\
\hline
\multicolumn{8}{|c|}{Pathology Classes} \\
\hline
Training sample & Acc. & R & R-OC & R-OS & P & P-OC & P-OS \\
\hline
1000 & 98.64 & 67.41 & 44.72 & 60.00 & 69.73 & 44.57 & 56.22 \\
3023 (all) & 99.54 & 90.74 & 86.94 & 80.67 & 88.45 & 86.35 & 78.72 \\
\hline
\end{tabular}
\end{center}
\label{tab:text_cnn_results_all}
\end{table}

\begin{table}
\caption{BLEU1-4 score comparisons on model in~\cite{shin2016learning} and our RNN0, RNN1 and RNN2 trained on all gold-standard MeSH annotations, and trained on 1000 gold-standard MeSH annotations+textCNN predictions.}
\begin{center}
\begin{tabular}{|l|c|c|c|}
\hline
& Train & Val & Test \\
\hline
Model & B-1/B-2/B-3/B-4 & B-1/B-2/B-3/B-4 & B-1/B-2/B-3/B-4 \\
\hline
Learning to read~\cite{shin2016learning} & \textbf{97.2} / \textbf{67.1} / 14.9 / 2.8 & \textbf{68.1} / 30.1 / 5.2 / 1.1 & \textbf{79.3} / 9.1 / 0.0 / 0.0\\
\hline
RNN0+vgg16+all & 8.8 / 1.8 / 0.7 / 0.2 & 7.8 / 2.6 / 1.1 / 1.4 & 6.9 / 2.3 / 0.7 / 0.1 \\
RNN0+resnet50+all & 16.5 / 8.6 / 4.6 / 2.4 & 16.7 / 8.7 / 3.9 / 1.0 & 18.8 / 10.4 / 3.8 / 1.9 \\
RNN1+resnet50+all &  77.9 / 45.6 / \textbf{29.4} / \textbf{18.9} & 65.7 / \textbf{51.6} / \textbf{30.2} / \textbf{17.1} & 66.7 / \textbf{47.1} / \textbf{26.8} / \textbf{15.9} \\
RNN2+resnet50+all & 74.1 / 42.0 / 26.8 / 17.3 & 63.2 / 47.5 / 27.3 / 15.8 & 63.2 / 43.9 / 25.5 / 13.6 \\
\hline
RNN0+resnet50+pred & 22.9 / 15.5 / 7.8 / 4.0 & 13.6 / 8.3 / 4.0 / 0.9 & 14.7 / 9.3 / 2.7 / 1.5 \\
RNN1+resnet50+pred & 73.6 / 50.0 / \textbf{30.9} / \textbf{17.8} & 41.5 / 29.7 / \textbf{15.9} / \textbf{7.2} & 41.6 / \textbf{28.2} / \textbf{13.2} / \textbf{8.1} \\
RNN2+resnet50+pred & 69.4 / 47.6 / 29.6 / 16.7 & 39.4 / 28.0 / 14.6 / 6.7 & 39.8 / 26.4 / 12.7 / 8.0 \\
\hline
\end{tabular}
\end{center}
\label{tab:seq_results_all}
\end{table}

\section{Results}
\subsubsection{Enriched Concept Extraction from Reports}
We evaluate the MeSH term prediction from the text reports by calculating the total binary accuracy (Acc), precision (P) and recall (R), and the mean-over-class (P-OC, R-OC) and mean-over-samples (P-OS, R-OS) precision and recall of the 102 classes. In addition, we report metrics of the `pathology' classes separately by manual allocation based on the definitions on the MeSH term online library~\cite{mesh}. Complete metrics are compared in Table~\ref{tab:text_cnn_results_all}.

\subsubsection{Report Generation from Images}
During inference, the first word is sampled from the LSTM, concatenated to the input, and used to predict consequent words. The quality of the generated reports was evaluated by measuring BLUE~\cite{papineni2002bleu} scores averaged over all the reports, which are a form of $n$-gram precision commonly used for evaluating image captioning as they maintain high correlation with human judgement. BLEU scores of RNN0, RNN1 and RNN2 trained on all gold-standard annotations and on the predictions made by the text CNN are presented in Table~\ref{tab:seq_results_all}. RNN0 is the same framework used in~\cite{shin2016learning}, however, they additionally train their model in a cascaded fashion which significantly improves the model's ability to predict the first word, but struggles to maintain visual correspondence in generating subsequent words, hence the steep reduction in higher $n$-gram precision. Additionally, cascaded models suffer from error propagation during test time, hence the poor performance on test data. RNN1 and RNN2 solve both problems by conditioning the word generation process on the images at every time-step and by being trained end-to-end, hence achieving higher $n$-gram scores on the test data. In addition, we have shown that we can achieve comparably high BLEU metrics when training on the predicted MeSH terms made by the text CNN.

\section{Conclusion}
We demonstrate how, given a small amount of manual annotations, clinically and visually-important concepts can be learned from raw textual radiology reports. We then demonstrate how these concepts can be used as radiological image annotations and used in an image-sequence learning model to auto-generate reports as part of a clinical decision support system.

\bibliographystyle{splncs04}
\bibliography{egbib}
\end{document}